\documentclass[conference]{IEEEtran}
\IEEEoverridecommandlockouts
\usepackage{cite}
\usepackage{amsmath,amssymb,amsfonts}
\usepackage{algorithmic}
\usepackage{graphicx}
\usepackage{textcomp}
\usepackage{xcolor}
\usepackage{url}
\usepackage{array}
\usepackage[hidelinks]{hyperref}
\def\BibTeX{{\rm B\kern-.05em{\sc i\kern-.025em b}\kern-.08em
    T\kern-.1667em\lower.7ex\hbox{E}\kern-.125emX}}
\begin{document}

\title{AntiCheatPT: A Transformer-Based Approach to Cheat Detection in Competitive Computer Games}

\author{\IEEEauthorblockN{Anonymous Authors}}

\author{\IEEEauthorblockN{1\textsuperscript{st} Mille Mei Zhen Loo}
 \IEEEauthorblockA{
\textit{The IT-University of Copenhagen}\\
Copenhagen, Denmark \\
milo@itu.dk}
 \and
 \IEEEauthorblockN{2\textsuperscript{nd} Gert Luẑkov}
 \IEEEauthorblockA{
 \textit{The IT-University of Copenhagen}\\
 Copenhagen, Denmark \\
 gelu@itu.dk}
 \and
 \IEEEauthorblockN{3\textsuperscript{rd} Paolo Burelli}
 \IEEEauthorblockA{\textit{brAIn lab - Culture, Play and AI Section} \\
 \textit{The IT-University of Copenhagen}\\
 Copenhagen, Denmark \\
 pabu@itu.dk}
 }

\maketitle

\begin{abstract}
Cheating in online video games compromises the integrity of gaming experiences. Anti-cheat systems, such as VAC (Valve Anti-Cheat), face significant challenges in keeping pace with evolving cheating methods without imposing invasive measures on users' systems. This paper presents AntiCheatPT\_256, a transformer-based machine learning model designed to detect cheating behaviour in Counter-Strike 2 using gameplay data. To support this, we introduce and publicly release CS2CD: A labelled dataset of 795 matches. Using this dataset, 90,707 context windows were created and subsequently augmented to address class imbalance. The transformer model, trained on these windows, achieved an accuracy of 89.17\% and an AUC of 93.36\% on an unaugmented test set. This approach emphasizes reproducibility and real-world applicability, offering a robust baseline for future research in data-driven cheat detection.
\end{abstract}


\section{Introduction}

Despite the success of online games, such as Counter-Strike 2 (CS2), many games suffer from malicious parties that reduce the enjoyment of honest players seeking a challenge. Games using Valve Anti-Cheat (VAC), such as Counter-Strike 2 and Team Fortress 2, are especially under scrutiny due to the poor performance of the anti-cheat measures. Furthermore, the low prices of cheats that can bypass VAC make cheating very accessible \cite{collins_anti-cheat_2024}.

The design of VAC, like that of many other anti-cheats, is intentionally obscure, despite security concerns \cite{newell_valve_2014}. Due to the closed nature of commercial anti-cheat systems, academic research in cheat detection and open-source alternatives remains limited. Although machine learning has been applied in this field, such research is often constrained by the availability of data~\cite{galli_cheating_2011,yeung_detecting_2006,kanervisto_gan-aimbots_2023,jonnalagadda_robust_2021,willman_machine_2020}. This scarcity of data poses a significant challenge, particularly for data-intensive deep learning models. 

To the best of our knowledge, there exists only one published paper that applies transformer models to behaviour based cheat detection in video games. The reported results seem promising, inferring that transformers can be effective in identifying patterns in player behaviour with an accuracy of 96.94\% and an AUC of 98.36\%. Unfortunately, this paper does not go into any architectural or implementation detail, nor does it provide access to the code and data used. Although the authors reference the original ``Attention Is All You Need'' paper~\cite{vaswani_attention_2023} as a basis for their model, no concrete architecture, hyperparameters, or preprocessing steps are disclosed, thereby making it impossible to assess and reproduce their results~\cite{tao_xai-driven_2020}.

The combination of limited data access and closed-source practices hinder the ability to build on prior research. This paper seeks to address that gap by developing and releasing a public labelled dataset and a transformer-based model for cheat detection using data extracted from CS2 gameplay.

\section{Dataset}

The Counter-Strike 2 (CS2) game was chosen due to its widespread cheating problem and a feature allowing players to download match replays in the form of DEM files. This contains detailed records of player actions, movement, and interactions during matches, enabling in-depth analysis of player behaviour. The existence of these files provided a unique opportunity to build a labelled dataset for training and evaluating machine learning models for cheat detection.

\subsection{Data collection}

While the log data for CS2 is not publicly available by default, we based the data collection on the data available from the match sharing service \href{https://csstats.gg/}{csstats.gg}. 

This site allows participants to upload their matches and displays the 50 most recent matches. Furthermore, the page indicates whether a user participating in a match had been VAC-banned~\cite{esl_gaming_online_csstats_nodate}. This information was used to create a script that scanned the site for publicly available match sharing codes, which can be used to download DEM files from Valve servers. The match sharing code, names of the VAC-banned users, and other match information were saved to a CSV file. This file was used to select a balanced set of matches with respect to the played map type. Unfortunately, this was not possible for matches where VAC-banned users were present, as there was a tendency for cheaters to prefer some maps over others. Hence, the data collection resulted in 795 matches, with 317 having at least one VAC-banned player and 478 having no VAC-banned players\footnote{Code regarding the data collection is available at: \url{https://github.com/Pinkvinus/CS2_demo_scraper}}.

\subsection{Data labelling}

The integrity of the labelling system is critical for the quality of the dataset. Taking a subsample ($n=50$) of the data without VAC-banned players present, the precision of the non-cheater label was 97. 2\%. Taking a subsample ($n=50$) of the set with at least one VAC-banned player showed that the precision of the VAC-ban in detecting cheaters was 92.6\% with a recall of 24.7\%. Due to the poor metrics observed for this subset, all 317 DEM files containing at least one VAC-banned player were manually reviewed.
The criteria for a player being labelled a cheater are that there was evidence of cheating beyond a reasonable doubt. This implies that, in the scenario where a player was performing well but did not display any behaviour of a cheater, the player would not be labelled as a cheater unless undeniable evidence was present. In the case of uncertainty, a player was labelled as not a cheater.
Whilst labelling the data, outliers were detected. Most notably, games where no rounds were played. Three of these outliers were subsequently removed from the dataset.

\subsection{Data Extraction and publication}

Demo files contain sensitive information. Therefore, the raw data could not be published directly. To allow publication, it was necessary to apply anonymisation procedures. Unfortunately, no publicly available software can perform this anonymisation on a DEM file directly, which requires the data to be extracted into another format before anonymisation can begin. Unfortunately, this means that matches within the dataset would not be replayable as they were no longer in the DEM file format.

For data extraction of DEM files, this project used the Python library demoparser2~\cite{laihoe_demoparser2_2024}. This parser is able to parse CS gameplay into two formats, ticks and events, using the functions \texttt{parse\_event()} and \texttt{parse\_ticks()}. Ticks are stored in a time-sequential Pandas DataFrame. Events are stored as a list of tuples containing strings and DataFrames. Using the Python type annotation the following is the data type returned by \texttt{parse\_event()}:

\begin{center}
    \texttt{list[tuple[str, pd.DataFrame]]}
\end{center}

For this dataset to be publishable, it was necessary to apply anonymisation procedures. We deemed there were two categories of data, omittable data and essential data. The omittable data would be removed from the dataset, and the essential data would undergo anonymisation procedures. Essential data only included IDs of the players, whereas the omittable data included names and cosmetic identifiers.

Notably, if all players from a single team leave, a single bot spawns in their place. The bot has no AI, remaining perfectly still in the spawn site of the empty team. Therefore, kills for a user with no ID can occur. These are bot kills.

\subsection{Data Publication}

The resulting dataset contains a total of 795 matches, 317 of which contain at least one cheater and 478 contain no cheater labels. Each match is represented by two files: a PARQUET file for tick data and a JSON file for event data. The published dataset is 48.9 GB and is published as the CS2CD dataset, short for Counter-Strike 2 Cheat Detection, on huggingface\footnote{The dataset is available at \url{https://doi.org/10.57967/hf/5315}.}.





\section{Cheat detection using Transformers}
\subsection{Context windows}

The data used as input for the transformer models in this paper is segmented into multivariate time series blocks (context windows), containing a sequence of tick vectors, where each tick vector is comprised of specific data points from a match within the dataset. The game CS2 updates it's state 64 times a second. A single context window is centred around a single kill, which contains data from two players: The attacker and the victim. For the AntiCheatPT\_256 model, a context window of 256 ticks, equivalent to 4 seconds, was selected. For this context window, 224 ticks were taken before the kill happened, and 32 after the kill happened. Each tick vector contained 44 data points. Therefore, the input to the AntiCheatPT\_256 model was a matrix of 256 rows by 44 columns. From the CS2CD dataset, 90,707 context windows have been extracted, with 18,423 being cheater kills and 72,284 being non-cheater kills.

\subsubsection{Data augmentation}

Data augmentation was performed on the attacker and victim's X, Y and Z coordinates. This is intended to prevent the models from overfitting on specific coordinates, where cheater or non-cheater kills happen, but rather to focus on a function of relative positioning between two players. Therefore, Gaussian noise was added in small amounts to the coordinates of the attacker and victim. Note that the exact same amount of random noise is added to the same coordinate of the attacker and victim, so that their relative positioning and distance from each other remain the same.

The cheater data was augmented more than the non-cheater data. Specifically, for every cheater data context window, three augmentations were made, meaning the amount of cheater context windows increased from 18,423 to 73,692. For every non-cheater context window, one augmentation was made, meaning the number of context windows with a non-cheater label increased from 72,284 to 144,568. The total number of context windows after the data augmentation was 218,260, resulting in a cheater context windows to non-cheater context windows ratio of approximately 1:2.

\subsubsection{Tick vector}

The first 24 data points in the vector describe the attacker's data or information about the kill within this context window, the first 17 of which describe the state of the attacker and the kill. The final attacker values are a one-hot encoding of the weapon group held by the attacker. Such weapon groups include knives, automatic rifles, semi-automatic rifles, pistols, grenades, submachine-guns, and shotguns. The following five values describe the state of the victim. Furthermore, noise data is included because it enhances the credibility of kills through smoke or walls, as it allows the attacking player to accurately infer the victim's position without direct visual confirmation. Finally, a one-hot encoding of the played map is added with 15 possible maps present, resulting in a tick vector of length 44\footnote{The context windows and further documentation are available at: \url{https://www.doi.org/10.57967/hf/5656}}.

\subsection{Transformer architecture}

The AntiCheatPT's model architecture is a transformer encoder based on the original implementation of the transformer by Vaswani et al. \cite{vaswani_attention_2023}. AntiCheatPT\_256 was created for the problem of fitting a function to behavioural cheat detection. Parameters such as feedforward layer dimension, number of attention heads, and number of transformer layers were all tested using various training hyperparameters to observe how the model's training converges. Model hyperparameters and training components can be seen in Table \ref{tab:training_comp}.

\begin{table}
    \centering
    \begin{tabular}{|m{14em}|m{14em}|}
        \hline
        Number of transformer encoder layers & 4  \\
        \hline
        Feedforward layer dimension & 176  \\
        \hline
        Attention heads & 1 \\
        \hline
        Loss function & Binary Cross Entropy (\texttt{BCEWithLogitLoss}) \\
        \hline
        Optimiser & AdamW (learning rate = $10^{-4}$)\\
        \hline
        Scheduler & StepLR (gamma = 0.5, step\_size = 10)\\
        \hline
        Batch size & 128 \\
        \hline
    \end{tabular}
    \vspace{0.5em}
    \caption{Training components for AntiCheatPT\_256}
    \label{tab:training_comp}
    \vspace{-8mm}
\end{table}

\subsubsection{Positional encoding}

A positional encoding was used for the input data to create positional dependency. The implemented positional encoding module precomputes a set of positional vectors using sine and cosine functions of varying frequencies, following the approach introduced in the original Transformer architecture by Vaswani et al.~\cite{vaswani_attention_2023}. For a given position $pos$, dimension $i$ and model feature dimension $d_{model}$, the encoding is defined as follows for even and odd $i$ respectively:
$$
    \texttt{PE}_{pos, 2i} =  sin \left( \frac{pos}{10000^{\frac{2i}{d_{model}}}}\right)
$$
$$
    \texttt{PE}_{pos, 2i+1} = cos \left( \frac{pos}{10000^{\frac{2i}{d_{model}}}} \right)
$$
The result of this contained values on the range from -1 to 1. Since the positional encoding would be added to a normalised numerical input in the range of 0 to 1, a scaling of the positional encoding was required. The following is the calculation of the final positional encoding:
$$
\texttt{PE}_{out} = \frac{\texttt{PE}_{in} + 1}{2} \cdot \texttt{PE\_scale}
$$
The \texttt{PE\_scale} value was chosen to be 0.1, which meant that all values within the positional encoding would be in the range from 0 to 0.1. 

\subsubsection{Model output}

For classification purposes, a classification token is prepended to the context window sequence. After passing through the transformer layers, this token is extracted as the aggregate representation of the sequence. The extracted token is then processed through two linear layers: the first transforms the 44-dimensional feature space into 128 dimensions, followed by a ReLU activation, and the second maps the 128-dimensional representation to a single output dimension. This output is subsequently passed through a sigmoid function to produce a probability score, where 1 indicates a cheating label and 0 indicates a non-cheating label. It is important to note that the sigmoid activation is not part of the model architecture itself, and must be applied externally. Therefore, the model outputs logits rather than probabilities.

\subsection{Model training}

\subsubsection{Data split}

The train, validation and test splits were respectively 70\%, 15\% and 15\% of the total dataset. Note that the data was split using file names from different matches as keys. These keys were then split into train, validation and test sets. From the keys, the context windows were retrieved. This way, no context windows from the same match would end up in different sets. That amounted to 159,592 (54,564 cheater and 105,028 not cheater) data points in the training set, 28,838 (10,168 cheater and 18,670 not cheater) in the validation set and 12,675 (2,240 cheater and 10,435 not cheater) in the test set. Note that the test set did not include any augmented data and was therefore smaller than the validation set. Thus, the test set contained a realistic scenario of inputs.

\subsubsection{Training setup}

The models were trained using early stopping to prevent overfitting. The random seed was set manually for reproducibility. The tested seeds were 41, 42 and 43. All seeds converged to the same minima, hence the seed 42 was chosen. The chosen batch size for training was 128. After each epoch, the validation set was tested. The validation metrics were saved along with the model parameters, optimiser parameters, scheduler parameters, epoch and average training loss as a PTH file\footnote{The model is available at: \url{https://www.doi.org/10.57967/hf/5653}}.

\section{Model Performance and Results}

The following metrics were obtained from test data, using a classification threshold of 0.7 or higher to identify cheaters, with the objective of minimising false positives:

\begin{table}[h]
    \centering
    \begin{tabular}{|l|c| c | c |}
        \hline
        \textbf{Metric} & \textbf{AntiCheatPT\_256} & \textbf{RNN}\cite{dunham_cheat_2020} & \textbf{Transformer}\cite{tao_xai-driven_2020}\\
        \hline
        Accuracy & 0.8917 & 0.8-0.9 & 0.9836 \\
        Precision & 0.8513 & - & - \\ 
        Recall & 0.6313 & - & -\\
        F1-score & 0.7250 & - & - \\
        Specificity & 0.9678 & - & - \\
        ROC AUC & 0.9336 & - & 0.9694\\
        \hline
    \end{tabular}
    \vspace{0.5em}
    \caption{Test metrics}
    \label{tab:test_metrics}
    \vspace{-4mm}
\end{table}

Compared to the claimed 80-90\% accuracy of the only other model that performs behavioural cheat detection on Counter-Strike game data specifically~\cite{dunham_cheat_2020}, the AntiCheatPT model shows similar results with an accuracy of 89.17\% as seen in table \ref{tab:test_metrics}. Unfortunately, no other metrics were given, other than the validation set accuracy; therefore, it is impossible to further compare the two models. Additionally, the only other model performing behavioural cheat detection using the transformer architecture had an accuracy of 98\% and an AUC of 97\%, thereby slightly outperforming the AntiCheatPT model, although no data or metrics regarding class imbalances were disclosed.

\begin{figure}[!t]
  \begin{minipage}[b]{0.49\linewidth}
    \centering
         \includegraphics[width=\textwidth]{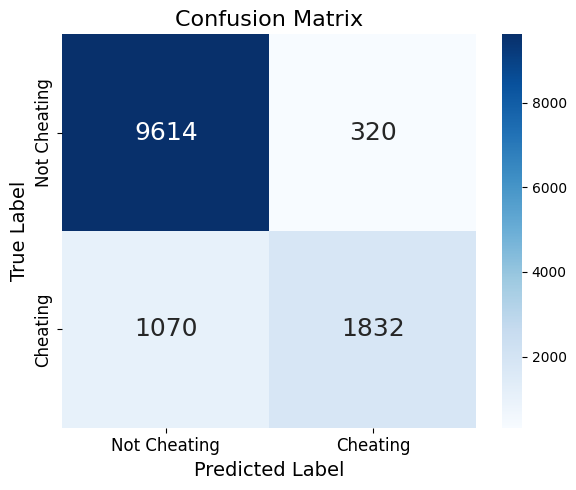}
         \caption{Confusion matrix}
         \label{fig:conf_matrix}
  \end{minipage}
  \hfill
  \begin{minipage}[b]{0.41\linewidth}
            \centering
         \includegraphics[width=\textwidth]{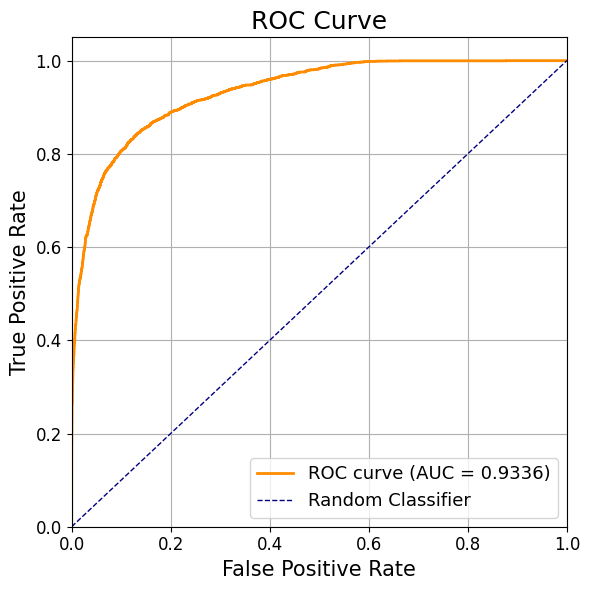}
         \caption{ROC}
         \label{fig:ROC}
    \label{fig:fig2}
  \end{minipage}
\end{figure}

\begin{figure}
    \centering
    \includegraphics[width=0.95\linewidth]{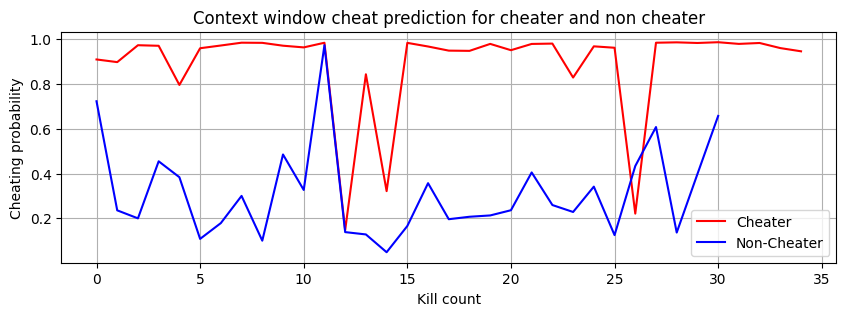}
    \caption{Cheating probability of kills for a cheater and non cheater}
    \label{fig:cheating_prob}
    \vspace{-4mm}
\end{figure}

The AntiCheatPT model shows good results in classifying the presence of cheating, with very few false positives present. Note, that when manually labelling cheaters, oftentimes a single kill is not enough to label a player as a cheater. Therefore, the rate of false negatives is not a big concern, however, a large false positive rate would be problematic. Therefore, in the case of real world applications, we recommend the evaluation of several kills rather than one. An example of a model inference for two players - a cheater and a non cheater in the same match, can be seen in figure \ref{fig:cheating_prob}. As can be seen on the figure, there are instances, where a cheating player gets a seemingly legitimate kill, and vice versa, where a non cheating player gets a kill, that raises the cheating probability. However, on average, the probability that a cheater is cheating is higher than that of the non cheating player.

Finally, using an RTX6000 48GB GPU, the training time of 4 epochs was $\sim$ 40 min and the test time was 42.45 s for 12675 data points resulting in an inference time of 3.35 ms per data point.
\section{Conclusion}

To address the lack of publicly available CS2 cheating datasets, this work fills a critical gap by offering a comprehensive, labelled resource for the research community. In total, the dataset contains 317 matches with manually labelled cheaters and 478 matches, with no VAC banned players present. Although a fully manually labelled dataset would be ideal, the process of manually labelling cheaters remains time-consuming.

Additionally, a machine learning solution using the transformer architecture was proposed, performing similarly to other models in the field, whilst being a fraction of the size of similar transformer models. All code and model weights are open-source, highlighting the reproducibility of this project as opposed to other projects using machine learning for cheat detection in video games. An area of development for the AntiCheatPT model is increasing the context window size while not overfitting to the training data, as that would increase the probability of cheating happening within a given context window.

Overall, this projects shows that an open-source reproducible method for behavioural cheat detection can be feasible for game companies that avoid developing kernel-level anti-cheats, such as Valve. This effectively shows that hardware-level cheats can be discovered through the use of a server-side machine learning anti-cheat system.

\bibliographystyle{IEEEtran}
\bibliography{references}

\end{document}